%% file: main.tex
\documentclass[letterpaper,10 pt,conference]{ieeeconf}
\usepackage{graphicx}
\usepackage{soul}
\newcommand{\AP}{{AP}}
\IEEEoverridecommandlockouts                              

\usepackage[linesnumbered,ruled,vlined]{algorithm2e}

\usepackage{xcolor}

\title{Optimal Control Synthesis with Relaxed Global Temporal Logic Specifications for Homogeneous Multi-robot Teams}

\author{Disha Kamale and Cristian-Ioan Vasile
\thanks{Disha Kamale and Cristian-Ioan Vasile are with the Mechanical Engineering and Mechanics Department,
Lehigh University, Bethlehem, PA 18015
{\tt\small \{ddk320, cvasile\}@lehigh.edu}}%
}

\usepackage{cite}
\usepackage{graphicx}
\usepackage{amsmath}
\usepackage{amssymb}
\usepackage{amsfonts}
\usepackage{mathtools}
\usepackage{tikz}
\usepackage{xcolor}
\usepackage{amsmath}
\usepackage{amsthm}
\usepackage{subcaption}
\usepackage{enumitem,amssymb}
\usepackage{pifont}
\usepackage{algorithm2e}

\newcommand{\disha}[1]{{\color{brown}#1}}
\newcommand{\True}{\mathsf{true}}

\newcommand{\PA}{{\mathcal{A}}}
    
\newtheorem{proposition}{Proposition}

\newtheorem{problem}{Problem}[section]
\newtheorem{example}{Example}[section]

\newcommand{\TS}{{\mathcal{T}}}
\SetCommentSty{mycommfont}

\makeatletter
\let\NAT@parse\undefined
\makeatother
\usepackage{hyperref}
\begin{document}
\maketitle

\setlength{\abovedisplayskip}{3pt}
\setlength{\belowdisplayskip}{3pt}
\setlength{\abovedisplayshortskip}{3pt}
\setlength{\belowdisplayshortskip}{3pt}

\setlength{\textfloatsep}{0.1cm}
\setlength{\floatsep}{0.1cm}

\input{def}


\begin{abstract}
In this work, we address the problem of control synthesis for a homogeneous team of robots given a global temporal logic specification and formal user preferences for relaxation in case of infeasibility.
The relaxation preferences are represented as a Weighted Finite-state Edit System and are used to compute a relaxed specification automaton that captures all allowable relaxations of the mission specification and their costs.
For synthesis, we introduce a Mixed Integer Linear Programming (MILP) formulation that combines the motion of the team of robots with the relaxed specification automaton.
Our approach combines automata-based and MILP-based methods and leverages the strengths of both approaches, while avoiding their shortcomings.
Specifically, the relaxed specification automaton explicitly accounts for the progress towards satisfaction, and the MILP-based optimization approach avoids the state-space explosion associated with explicit product-automata construction, thereby efficiently solving the problem.
The case studies highlight the efficiency of the proposed approach.
\end{abstract}

\section{Introduction}
One challenge in planning with complex temporal logic specifications in real-world applications concerns dealing with infeasible missions due to perhaps one small part of the requirements.
Recent advances in high-level planning using formal specifications have explored the idea of minimally relaxing the specification to ensure satisfaction.
Moreover, the problem is compounded when employing teams of robots due to combinatorial explosion of the problem search space.

Traditionally, the control synthesis problem has been addressed using automata-theoretic approaches owing to their rich expressivity.
In~\cite{fainekos2011revising}, the authors introduced the problem of \textit{minimal revision} for Buchi automata obtained from Linear Temporal Logic (LTL) specifications by quantifying the closeness between the automata.
These revisions, however, are challenging to interpret.
Moreover, the problem is shown to be NP-complete~\cite{kim2012approximate, kim2015minimal}.
Another automata-based notion of minimum revision was proposed in~\cite{lahijanian2016specification}.
The problem of \textit{minimum violation} based on preference-based violation was considered in~\cite{tuumova2013minimum,wongpiromsarn2021minimum, vasile2017minimum, lindemann2019control}. In~\cite{vasile2014automata,vasile2017time}, temporal relaxations was defined for timed specifications using a parameterized abstraction. The problem of \textit{partial satisfaction} was tackled in~\cite{cardona2022partial, cardona2023preferences} for Signal Temporal Logic and in~\cite{lacerda2015optimal} for co-safe LTL specifications. Similar problems were tackled in~\cite{rahmani2020you, ahlberg2019human, guo2018probabilistic} via hard and soft constraints. In~\cite{kamale2021automata}, we introduced an automata-based approach that combines various existing notions of relaxation into a unified framework and allows for complex word-word relaxation. Although the automata-based approaches are highly expressive, they cannot be readily used for multi-robot teams.

On the other hand, a rich body of work exists in planning for large robot teams given temporal logic specifications~\cite{kantaros2020stylus, kantaros2018sampling, sun2022multi,cardona2023temporal, leahy2021scalable, leahy2022fast}. In~\cite{kantaros2020stylus, kantaros2018sampling}, the authors proposed an informed sampling-based approach for the satisfaction of LTL specifications.  With the advent of off-the-shelf solvers~\cite{gurobi}, Mixed-Integer Linear Programming (MILP) formulations  have been utilized for large-scale planning problems~\cite{sun2022multi,cardona2023temporal, leahy2021scalable, leahy2022fast} 
These approaches do not consider for relaxation of the specifications if infeasibilities arise.

In this work, we consider the problem of planning for a team of homogeneous robots tasked with a global temporal logic specification and with formal user preferences.
We consider complex temporal logic goals with explicit timing constraints expressed as Time Window Temporal Logic (TWTL) formulae~\cite{vasile2014automata}. 
The user preferences are a set of complex rules for relaxations that are formally represented using a Weighted-Finite State Edit System (WFSE)~\cite{kamale2021automata}.
Using the automata corresponding to the mission and preferences, we construct a relaxed specification automaton that captures the original task alongside all permissible relaxations and their penalties.
To avoid the product of motion models which leads to exponential runtime with respect to the number of robots, we formulate an MILP over the transitions of the motion and relaxed specification models.
The encoding efficiently finds the globally optimal solution, bringing together the strengths of automata-based and optimization-based methods. 

Our approach differs from the existing MILP-based and relaxation techniques in the following aspects. The MILP formulations defined in~\cite{leahy2021scalable, cardona2022partial, cardona2023preferences} optimize for robustness of satisfaction, whereas we are interested in optimally relaxing a specification by modifying it using given user-defined rules.
As opposed to spatial preferences as considered in~\cite{karlsson2020sampling}, we consider complex preferences of revisions.
In~\cite{peterson2021distributed}, the problem of distributed control synthesis for individual TWTL specifications allowing finite temporal relaxations is considered; whereas we consider planning for a global TL task with word-to-word revisions. Finally, the main difference as compared to our previous work~\cite{kamale2021automata} is that instead of considering a general problem of optimal relaxation over a 3-way product automaton for a single robot, we use an MILP formulation that simultaneously accounts for motion and specification relaxation for teams of robots.

The main contributions of this work are 1) We pose the problem of optimal planning with relaxation for a homogeneous robot team subject to a global temporal logic specification. 2) We define a \textit{relaxed specification automaton} that represents the given global specifications with all allowable revisions. 3) We propose a framework that combines automata and MILP-based approaches to address the minimal relaxation planning problem. The relaxed specification automaton explicitly accounts for the progress towards satisfaction, while the MILP formulation helps to avoid scalability issues, thereby bringing together the strengths of both approaches while avoiding their limitations. 
4) We demonstrate the functionality of the proposed framework with the help of case studies and provide a runtime analysis. We compare the proposed MILP formulation with a baseline for their efficiency and expressivity.

\noindent\textbf{Notation.}
The set of real, integer, and binary numbers are represented by $\mathbb{R}$, $\mathbb{Z}$, and $\mathbb{B}$.
The set of integer numbers greater or equal to $a$ is defined by $\mathbb{Z}_{\geq a}$.
The integer range from $a$ to $b$ is denoted by $\range{a,b}$.
For a set $X$, $2^X$ and $|X|$ represent its power set and cardinality, respectively. For an alphabet $\Sigma$, a language of all finite words over $\Sigma$ is denoted by $\Sigma^*$.  We refer to a directed acyclic graph as a DAG. 

\section{Problem Setup}

In this section, we introduce the problem of optimal planning for a global temporal logic goal with formal preferences for relaxations. Using a finite system abstraction for the environment and automata to capture the progress towards goal and preferences, we define a cost function that combines the costs incurred for motion and relaxations.
Next, we pose the formal problem definition. 

\subsection{Robot and Environment Models}


We consider a team of $N$ identical robots, denoted by $\mathcal{R} = \{r_1, r_2, \ldots, r_{N}\}$, deployed in an environment to perform a global task. Each robot can move in the environment deterministically.
We assume that the robots move in the environment synchronously.
This is achieved either by assuming robots have access to a global clock, synchronize via inter-robot communication, or receive and execute commands from a global controller.

The robots' motion in the environment is abstracted as a transition system $\TS$ defined as follows:   
\begin{definition}[Transition System]
A weighted transition system (TS) is a tuple
$\TS = (X, \mathcal{R}, \CA{I}^\TS, \delta_\TS, \AP, h, w_\TS)$, where
$X$ is a finite set of states and corresponds to regions in the environment;
$\CA{I}^\TS : \mathcal{R} \to X$ defines the robots' initial states;
$\delta_\TS \subseteq X \times X$ is a set of transitions that capture the set of allowable movements in the environment;
$\AP$ is a set of labels (atomic propositions);
$h : X \to \spow{\AP}$ is a labeling function;
$w_\TS : \delta_\TS \to \BB{R}_{\geq 0}$ is a weight function that denotes control effort incurred in traversing $(x, x') \in \delta_\TS$.
%
\end{definition}

The states $X$ of the environment represent locations connected by directed transitions $\delta_\TS$.
We use self-loop transitions $(x, x) \in \delta_\TS$ to model robots stationary at state $x$.
All transitions are assumed without loss of generality to have duration 1 that corresponds to a fixed time discretization step $\Delta t$.
As a robot $r_i \in \mathcal{R}$ moves through the environment, it generates a sequence of states  $\BF{x}_{r_i} = x_{r_i, 0} x_{r_i,1} \ldots$, referred to as a \emph{trajectory} (or run) of a robot, such that $(x_{r_i,k}, x_{r_i, k+1}) \in \delta_\TS$ for all $k \in \mathbb{Z}_{\geq 0}, i \in [[1,N]]\textit{}$ and $x_{r_i, 0} = \CA{I}^\TS(r_i)$.
An atomic proposition $\pi \in AP$ is said to be $\True$ when at least one robot is at a state $x$ labeled with $\pi$, i.e., $x \in h^{-1}(\pi)$.
Overall, the observations at states are given by the entire team of robots $\mathcal{R}$.
Thus, we introduce the team state and output trajectories.

\begin{definition}
The team trajectory $\BF{\Tilde{x}}$ is a sequence of vectors of all states occupied by robots at each time instance, i.e.,  $\BF{\Tilde{x}}(k) = [\BF{x}_{r_i}(k)]_{r_i \in \mathcal{R}} \in X^{|\mathcal{R}|}$ for a global time index $k$. The corresponding output word is $\BF{o}=o_0 o_1 \ldots$ with $o_{k} = \bigcup_{r_i \in \mathcal{R}} h(x_{r_i, k})$.
The {\em (generated) language} of $\TS$ is
the set of all team output words, denoted by $\CA{L}(\TS)$.
\end{definition}

The control cost of a transition from state $x$ to $x'$ is captured by $w_\TS(x, x')$.
When robots are stationary, they incur zero control cost, i.e., $w_\TS(x, x) = 0$ for all $x \in X$.
The \textbf{\textit{control cost}} for a team trajectory $\BF{\Tilde{x}}$ is defined as 
    $\hat{J}_{control}(\BF{\Tilde{x}}) = \sum_{r_i \in \mathcal{R}} \sum_{k \in \vert \BF{\Tilde{x}} \vert -1 } w_\TS(x_{r_i, k} , x_{r_i, k+1})$.

This work focuses on high-level decision-making for a complex global task with relaxations, we assume that appropriate low-level controllers are provided that facilitate the execution of the high-level plans.

\subsection{Task Specification}

    

Owing to its expressivity and conciseness, the global task is given as a Time Window Temporal Logic (TWTL) formula, introduced in~\cite{vasile2014automata}. 
A TWTL formula is defined over a set of atomic propositions and its syntax expressed in Bakus-Naur form is 
\begin{equation*}
\label{eq:logic-def}
\phi :: = H^d s \, |
\, \phi_1 \land \phi_2 \, | \, \phi_1 \lor \phi_2 \, | \, \phi_1 \cdot \phi_2 \, | \, [\phi_1]^{[a, b]}
\end{equation*}

\noindent where $s$ is either the ``true'' constant $\True$
or an atomic proposition in $\AP$;
$\land$ and $\lor$ are the conjunction and disjunction
Boolean operators, respectively; $\cdot$ is the concatenation operator;
$H^d$ with $d \in \BB{Z}_{\geq 0}$ is the {\em hold} operator;
and $[\ ]^{[a,b]}$
is the {\em within} operator, $a, b \in \BB{Z}_{\geq 0}$ and $a \leq b$.
For a thorough description of the syntax and semantics of TWTL, we refer the reader to~\cite{vasile2017time, vasile2014automata}. 

For simplicity, we disallow negations of atomic propositions, i.e., $H^d \neg s$,
which require additional bookkeeping in our solution in Sec.~\ref{sec:approach}.
We leave this for future work.
However, we still require the formulas without ambiguous concatenation, see~\cite{vasile2017time}.

\subsection{User preferences} 
The user preferences govern the modifications to the original mission specification $\phi$ if infeasibilities arise.
%
Formally, let $L$ denote a language over an alphabet $2^\AP$.
A \emph{user task preference} is a pair $(\Pref, w_\Pref)$,
where $\Pref\subseteq L\times(2^\AP)^*$ is a relation specifying how words in $L$ can be transformed to words from $(2^\AP)^*$;
and $w_\Pref \colon \Pref \to \BB{R}_{\geq 0}$ represents
the cost associated with the word transformations.
Alternatively, preferences are given by a set of word-rewrite rules with associated weights.
We denote a weighted word-rewrite rule as $\rho = \BS{\sigma} \mapsto^{w_{\BS{\sigma},\BS{\sigma}'}} \BS{\sigma}'$ where $\BS{\sigma}_{pre}\BS{\sigma}\BS{\sigma}_{suf} \in L, \BS{\sigma}_{pre}\BS{\sigma}'\BS{\sigma}_{suf} \in (2^\AP)^*$ for some prefix and suffix words $\BS{\sigma}_{pre}$ and $\BS{\sigma}_{suf}$, respectively.
To allow satisfaction without relaxation, we require that $(\BS{\sigma}, \BS{\sigma}) \in R$ with $w_R(\BS{\sigma}, \BS{\sigma}) = 0$ for all $\BS{\sigma} \in L$.
The example below illustrates the definition.

\begin{example}
 Consider the specification ''Visit A and B for 2 and 3 time units within the first 5 minutes followed by C for 1 time unit within the next 1 minute." Expressed as a TWTL formula, we have $\phi = [H^2 A \land H^3 B]^{[0,5]} \cdot [H^1 C]^{[0,1]}$. The relaxation preferences are given as follows - 1) $C$ can be substituted by a visit to region $D$ with a penalty of 3. 2) $A$ can be substituted by simultaneously visiting $B$ and $D$ with a penalty of 2. Formally, $\rho_1 = C \mapsto^3 D, \rho_2 = A \mapsto^2 BD$. 
\end{example}



The multi-robot control synthesis problem is as follows.

\begin{problem}[Optimal Control Synthesis with Relaxation]
\label{prob:relaxed planning}
Given a team $\mathcal{R}$, a discrete abstraction of the environment $\TS$, a global task specification expressed as a TWTL formula $\phi$, and user preferences for specification relaxation as $(R, w_R)$, the problem of optimal control synthesis with minimal relaxations corresponds to finding a team trajectory $\BF{\Tilde{x}}$ that satisfies a potentially relaxed version of the mission $\phi$ while minimizing the control and relaxation costs. 
Formally, 
\begin{align*}
     \min_{\tilde{\BF{x}}} \hat{J}(\BF{\Tilde{x}}) = \hat{J}_{control}(\BF{\Tilde{x}}) + \lambda \cdot w_R(\BF{o}, \mathbf{o}^{relax}),\\
     \text{s.t. } \BF{o}^{relax} = h(\BF{\Tilde{x}}), \quad (\BF{o}, \BF{o}^{relax}) \in R, \quad \BF{o} \models \phi,
\end{align*}
where $\lambda$ is a blending parameter, $\BF{o}^{relax} = h(\BF{\Tilde{x}})$ is the output word of the $\TS$, and $\BF{o} \in (2^\AP)^*$ is a word satisfying $\phi$.
\end{problem}

\section{Approach}
\label{sec:approach}

In this section, we present a solution to Problem~\ref{prob:relaxed planning}.
Our approach combines the expressivity of automata with the efficiency of optimization-based approaches.
We proceed by translating the specification and user preferences to automata and then constructing a product automaton that encapsulates the original specification and all its permissible revisions.
We present an algorithm to compute this relaxed specification product which is defined using a set of input-output symbol pairs in the WFSE.
We then present the main contribution of this work -- an MILP formulation that combines the motion model with the relaxed specification automaton as a solution to Problem~\ref{prob:relaxed planning}.
Critically, we avoid the construction of product transition systems, a common in automata-based methods~\cite{vasile2014automata, ulusoy2011optimal,nikou2016cooperative}.

\subsection{Specification and User Preferences Automata}

In order to solve the problem of control synthesis for a given mission $\phi$, a deterministic finite state automaton, (defined below), is recursively constructed from $\phi$ using the procedure discussed in~\cite{vasile2017time}. 

\begin{definition}[Deterministic Finite State Automaton]
\label{def:dfa}
A deterministic finite state automaton (DFA) is a tuple
$\FA = (S_\FA, s_0^\FA, \Sigma, \delta_\FA, F_\FA)$, where
$S_\FA$ is a finite set of states;
$s^\FA_0 \in S_\FA$ is the initial state;
$\Sigma$ is the input alphabet;
$\delta_\FA : S_\FA \times \Sigma \to S_\FA$ is the transition function;
$F_\FA \subseteq S_\FA$ is the set of accepting states.
%
\end{definition}
A trajectory of the DFA $\BF{s} = s_0 s_1 \ldots s_{n+1}$ is generated by
a finite sequence of symbols $\BS{\sigma} = \sigma_0 \sigma_1 \ldots \sigma_n$
if $s_0 = s^\FA_0$ is the initial state of $\FA$ and
$s_{k+1} = \delta_\FA(s_k, \sigma_k)$
for all $k \geq 0$.
A finite input word $\BS{\sigma}$ over $\Sigma$ is said to be accepted
by a finite state automaton $\FA$ if the trajectory of $\FA$ generated
by $\BS{\sigma}$ ends in a state belonging to the set of accepting states, i.e., $F_\FA$.
The {\em (accepted) language} of a DFA $\FA$ is
the set of accepted input words denoted by $\CA{L}(\FA)$.

In~\cite{kamale2021automata}, we showed that formal user preferences can be represented as a special class of automata called Weighted-Finite State Edit System (WFSE), defined as follows.
\begin{definition}[Weighted Finite State Edit System]
\label{def:wfses}
A weighted finite state edit system (WFSE)
is a weighted DFA $\ES = (Z_\ES, z^\ES_0, \Sigma_\ES, \delta_\ES, F_\ES, w_\ES)$, where 
$\Sigma_\ES = 2^\AP \times 2^\AP$,
and $w_\ES\colon \delta_\ES \to \BB{R}_{\geq 0}$ is the transition weight function.
\end{definition}

The alphabet $\Sigma_\ES$ captures the substitution operations.
A weighted transition $z' = \delta_\ES(z,(\sigma, \sigma'))$
has input, output symbols $\sigma$ and $\sigma'$.
Given a word $\vec{\BS{\sigma}} = (\sigma_0, \sigma'_0) (\sigma_1, \sigma'_1) \ldots (\sigma_r, \sigma'_r) \in \CA{L}(\ES)$, {\color{black}$r=|\vec{\BS{\sigma}}|-1$,}
we refer to $\BS{\sigma}= \sigma_0 \sigma_1\ldots \sigma_r \in (2^\AP)^*$ as \textit{input word}
and to $\BS{\sigma}'=\sigma'_0 \sigma'_1\ldots \sigma'_r \in (2^\AP)^*$ 
as the \textit{output word}.
Moreover, we say that $\ES$ transforms $\BS{\sigma}$ into $\BS{\sigma}'$.

\noindent
\textit{Remark:} In this work, we are particularly interested in addressing the minimum revision problem with complex preferences. Thus, as opposed to a more general alphabet used for WFSEs in~\cite{kamale2021automata}, we drop the inclusion of the \textit{empty symbol} $\epsilon$ in the current work. Investigating the problems of minimum violation, insertion and temporal relaxation are topics for future research. 

The \textbf{\textit{relaxation cost}} for transforming  $\BS{\sigma}$ to $\BS{\sigma}'$ is given by 
    $\hat{J}_{revision}(\vec{\BS{\sigma}}) = \sum_{l \in  \range{0, |\vec{\BS{\sigma}}|}} w_\ES(\sigma_l, \sigma_l')$.

\subsection{Relaxed Specification Product Automaton}

We capture all relaxed versions of the TWTL specification $\phi$ via a product non-deterministic finite state automaton between the DFA $\FA$ and the WFSE $\ES$.

\begin{definition}[Relaxed Specification Automaton]
    Given a specification DFA $\FA = (S_\FA, s_0^\FA, \Sigma, \delta_\FA, F_\FA)$, the user task preferences represented as a WFSE $\ES = (Z_\ES, z^\ES_0, \Sigma_\ES, \delta_\ES, F_\ES, w_\ES)$, the relaxed specification automaton is 
    a tuple $\PA = (Q_{\PA}, q_{\PA}^0, \Sigma_\PA, \delta_{\PA}, F_{\mathcal{A}}, w_\PA)$, where $Q_\PA = Z_\ES \times S_\FA$ represents the state space; $q^\PA_0 = (z^\ES_0, s_0^\FA)$ is the initial state; 
        $\Sigma_\PA = \Sigma_\FA$ denotes the alphabet; 
        $\delta_\PA \subseteq Q_\PA \times \Sigma_\PA \times Q_\PA$ is a transition relation; 
        $F_\PA = F_\ES \times F_\FA \subseteq Q_\PA$ represents the set of final (accepting) states;
        $w_\PA : \delta_\PA \rightarrow \mathbb{R}_{\geq 0}$  is the weight function.

    
\end{definition}

The procedure for constructing $\PA$ is outlined in Alg.~\ref{alg:PA}. All components of the product are initialized, and the initial state is added to the stack of nodes to be processed (lines 1-3).
For a given input symbol $\sigma$, a transition $((z,s), \sigma, (z', s')) \in \delta_\PA$ if and only if there exists $\sigma'$ such that $z' \in \delta_\ES(z, (\sigma, \sigma'))$ and $s' = \delta_\FA(s, \sigma')$.
Given the current states $(z, s)$, we iterate over each outgoing edge of $z$ in $\ES$ (line 5)
with input-output symbol pair $(\sigma, \sigma')$.
The next state in the DFA $\FA$ is determined given the current state $s$ and the output symbol $\sigma'$ (line 6).
The states and edges are added to the product and to $F_\PA$, if applicable. 
The weight function is defined as $w_\PA((z,s), \sigma, (z', s')) = w_\ES(z, (\sigma, \sigma'), z')$.
The process repeats until no states $(z, s)$ are left to process in the stack.
Thus, the relaxed specification automaton simultaneously captures the original specification, and user preferences for relaxations along with their penalties. 

\begin{algorithm}
\caption{\textit{construct\_{$\PA$}}()}
\label{alg:PA}
\scriptsize{
\KwIn{$\ES, \FA$}
\KwResult{$\PA = (Q_{\PA}, q^{\PA}_0, \Sigma_\PA, \delta_{\PA}, F_{\PA}, w_\PA)$}
\DontPrintSemicolon
\BlankLine
$\Sigma_\PA \gets \Sigma_\FA$,
$q_\PA^0 \gets (z^\ES_0, s_0^\FA)$, $Q_{\PA} \gets \{q_\PA^0\}$, $\delta_\PA \gets \emptyset$, $F_\PA \gets \emptyset$\; 

$stack \gets Q_\PA$ \; 
\While{$stack \neq \emptyset$}{
    $(z, s) \gets stack.pop()$ \;
    \ForAll{$(z, (\sigma, \sigma'), z') \in \delta_\ES$}{
        $s' \gets \delta_\FA(s,\sigma')$ \;
        \If{$(z',s') \notin Q_\PA$}{        
                $Q_\PA \gets Q_\PA \cup \{(z', s')\}$ \;
                $stack.push((z', s'))$ \;
                \lIf{$z' \in F_\ES$ and $s' \in F_\FA$}{
                    $F_\PA \gets F_\PA \cup \{(z', s')\}$
                }
            }
            $\delta_\PA \gets \delta_\PA \cup \{((z, s), \sigma, (z', s')))\}$\;
            $w_\PA((z, s), \sigma, (z', s'))) \gets w_\ES(z, (\sigma, \sigma'), z')$
        }
    }
} 
\end{algorithm}

The relaxed specification automaton $\PA$ is a non-deterministic model due to the potentially multiple ways to translate an input symbol $\sigma$ into an output $\sigma'$, see lines 5, 6 and 11 in Alg.~\ref{alg:PA}.
Semantically, this means that there are potentially multiple ways to relax the mission specification, as illustrated in the example below.

\begin{example}
    Consider user preferences: $A \mapsto^{p1} B$, $AB \mapsto^{p2} CD$, and $AC \mapsto^{p3} BD$. The same input symbol $A$ may result in different output symbols leading to non-determinism in the relaxed specification automaton. 
\end{example}

\begin{proposition}
\label{prop:DAG}
$\PA$ is a directed acyclic graph (DAG).
\end{proposition}
This follows from the fact that the recursive translation of $\phi$ into $\FA$ as proposed in \cite{vasile2017time} results in a DAG. In general, if $\FA$ is a DAG, the resulting product $\PA$ is also a DAG. 
Proposition~\ref{prop:DAG} offers a key insight into our MILP formulation since the product space assumes a directionality that can be leveraged for flow-based encoding. 
  
\subsection{Mixed Integer Linear Programming (MILP$_\ES$)}

Given $\TS$ and $\PA$, we now pose Problem~\ref{prob:relaxed planning} as an instance of an MILP. Towards this, we make the following assumptions: 1) The specification $\phi$ does not contain any negations; 2) All transitions in $\TS$ take unit duration for traversal. 
These assumptions are not restrictive since, in the case of assumption (1), the forbidden states can be eliminated during environment abstraction. 
Regarding assumption (2), auxiliary states and transitions can be used in case of longer durations.
The MILP encoding uses the observation that we do not need to track robots in $\TS$.
It suffices to compute the number of robots at each state and transition over time.

Let $n(x)$ denote the initial number of robots at the state $x \in X$ such that $\sum_{x\in X} n(x) = N$, and $X^\TS_0 = (\CA{I}^\TS)^{-1}(\CA{R})$ the initial set of states occupied by robots.
We define integer variables $z_{x, x', q, q'} \in \range{0,N}$ to denote the number of robots traversing edge $(x,x')$ in $ \delta_\TS$ and $(q,q')$ in $\delta_\PA$.
Similarly, the integer variables $z_{x, q}$ correspond to the number of robots present at state $x\in X$ and state $q \in Q_\PA$. 

\subsubsection{Initialization Constraints}
Let the set of atomic propositions satisfied at deployment be $\AP_0 = h(X^\TS_0)$.
We introduce a virtual node $\bowtie$ that is connected to all nodes in $X_0^\TS$ with only outgoing edges, i.e., we add $\{(\bowtie, x) \mid x \in X^\TS_0\}$ of weight 0.
The set of states in $\PA$ that can be reached due to the satisfaction of the atomic propositions in $AP_0$ is given by $Q_0 = \bigcup_{\sigma \in 2^{\AP_0}} \delta_\FA(q_0, \sigma)$. \\
\begin{equation}
\label{eq:initialization}
    \sum_{q \in Q_0} z_{{\bowtie}, x, q_0, q} = n(x), \forall x \in X_0^\TS, 
\end{equation}

\subsubsection{Flow Constraints}
Since the overall number of robots should remain constant throughout the mission, the flow constraints require that the total number of robots entering a region must be equal to the total number of robots leaving that region. Let $\hat{Q}_\PA = Q \setminus F_\PA \setminus \{q^\PA_0\}$.
The set of predecessors of $q \in \hat{Q}_\PA$ is $\mathcal{N}^-_\PA(q) = \{q' \mid \exists \sigma \in 2^\AP, q \in \delta_\PA(q', \sigma)\}$.
Similarly, the set of successors is $\mathcal{N}^+_\FA(q) = \{q'\mid \exists \sigma \in 2^\AP, q' \in \delta_\PA(q, \sigma)\}$. Using these, the flow constraints are encoded as follows. 
{\small\begin{equation}
\label{eq:flow-constraint}
    \sum_{(x', x) \in \delta_\TS}\sum_{q'\in \mathcal{N}^-_\PA(q)} z_{x', x, q', q} = \sum_{(x, x') \in \delta_\TS}\sum_{q'\in \mathcal{N}^+_\PA(q)} z_{x, x', q, q'}.
\end{equation}}
 
\subsubsection{State-transition Constraints}
The number of robots at a state $x$ in the environment and $q$ in the specification is given by the outflow from~\eqref{eq:flow-constraint}, except for sink nodes with $q\in F_\PA$.
For all states $x \in X$ and $q \in Q$, 
\begin{equation}
\label{eq:state-transition}
    z_{x, q} = \begin{cases}
        \sum_{(x, x') \in \delta_\TS}\sum_{q'\in \mathcal{N}^+_\PA(q)} z_{x, x', q, q'}, & q \notin F_\PA\\
        \sum_{(x', x) \in \delta_\TS}\sum_{q'\in \mathcal{N}^-_\PA(q)} z_{x', x, q', q}, & q \in F_\PA
    \end{cases}.
\end{equation}

\subsubsection{Synchronization Constraints}
Since, at any instance, the entire team is in a single state of $\PA$, we synchronize the state transition in $\PA$ across all robots.
Let $\xi_{q, q'} \in \mathbb{B}$ denote whether a transition $(q, q') \in \delta_\PA$ was taken.
For $(q, q') \in \delta_\PA$, 
\begin{align}
  \sum_{q' \in \mathcal{N}^+_\PA(q)} \xi_{q, q'} \leq 1,    \label{eq:unique-next-state}\\
    \sum_{(x,x') \in \delta_\TS} z_{x,x', q, q'} \leq N \cdot \xi_{q, q'}. \label{eq:sync-transitions}
\end{align}
Since $\PA$ is a DAG, each edge in $\delta_\PA$ is traversed at most once. This is enforced by constraints~\eqref{eq:unique-next-state} and ~\eqref{eq:sync-transitions}, with the latter imposing the requirement on the robot flow.

\subsubsection{Guard Satisfaction Constraints}
%
The progress towards satisfaction is encoded as follows.
We process all transitions $(q, \sigma, q') \in \delta_\PA$ between $q$ and $q'$
together based on Boolean function representation $g_{q, q'}$ called a \emph{guard}.
Formally, $g_{q, q'}(\sigma)$ is true if and only if $(q, \sigma, q') \in \delta_\PA$.
We assume that the guards are in disjunction normal form (DNF), i.e., $g = \bigvee_{i=1}^{m^+} \eta_i$,
where each term $\eta_i = \bigwedge_{\ell=1}^{m^*} \pi_\ell $, $\pi_\ell \in AP$.
%
Let $Terms_\PA(q,q') = \{\eta_i\}_{i=1}^{m^+}$.
Let $\xi_{q', \eta} \in [0, 1]$ indicate whether the term $\eta$ of the guards leading to state $q'$ in $\PA$ is satisfied.
In other words, $\eta$ is a set of APs associated with the guard such that their simultaneous satisfaction enables the transition to $q'$.
Similarly, we define $\xi_{q', \pi} \in [0, 1]$ indicating whether the AP $\pi$ is satisfied for a guard leading to state $q'$ in $\PA$.
Note that $\eta$ and $\pi$ may be common across multiple incoming transitions to $q'$.
With this, the guard satisfaction constraints are as follows. 
\begin{align}
    \xi_{q, q'} &\leq \sum_{\eta \in Terms_\PA(q, q')} \xi_{q', \eta}, \label{eq:guard-sat}\\
    \xi_{q', \eta} &\leq \xi_{q', \pi}, \, \forall \pi \in \eta, \, \eta \in Terms_\PA(q, q'), \label{eq:guard-symbol-sat}\\
    \xi_{q', \pi} & \leq \sum_{x \in h^{-1}(\pi)} z_{x, q'}, \label{eq:guard-ap-sat}
\end{align}
$\forall (q, q') \in \delta_\PA$. Constraint~\eqref{eq:guard-sat} ensures that at least one of the terms associated with the guard of $(q,q')$ is satisfied. The satisfaction of all atomic propositions corresponding to a term is ensured by~\eqref{eq:guard-ap-sat}.
Finally, the motion in $\TS$ is connected with guard satisfaction in $\PA$ via $\xi_{q', \pi}$ which can be 1  only if there are robots at states labeled with AP $\pi$.

\subsubsection{Final state} 
The constraint to ensure that the final state in $\PA$ is reached, i.e., a relaxation of $\phi$ is satisfied, is 
\begin{equation}
\label{eq:acceptance-condition}
    \sum_{x \in X} \sum_{q \in F_\PA} z_{x, q} > 0.
\end{equation}

\subsubsection{Cost function}
Finally, we define the objective as follows 
$\hat{J} = \hat{J}_{control} + \lambda \hat{J}_{revision}$, $\lambda > 0$ 
\begin{align}
    \hat{J}_{control} &= \sum_{(x, x') \in \delta_\TS} \sum_{(q, q') \in \delta_\PA} w_\TS(x, x') z_{x, x', q, q'} \label{eq:cost-travel}\\
    \hat{J}_{revision} &= \sum_{(x, x') \in \delta_\TS} \sum_{(q, q') \in \delta_\PA} w_\PA(q, q') z_{x, x', q, q'} \label{eq:cost-relax}
\end{align}

\section{Case studies}

\begin{figure}
    \centering
    \includegraphics[width=.6\columnwidth]{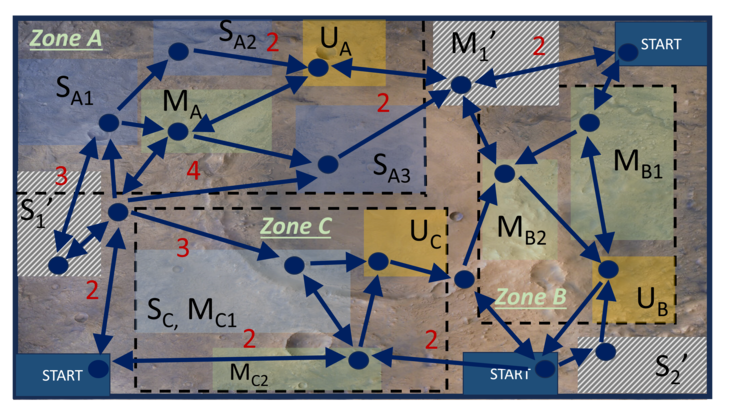}
    \caption{Transition system for Case Study 1}
    \label{fig:cs1_ts}
\end{figure}

In this section, we show the functionality of our approach, a comparison with a baseline, and a runtime analysis of MILP$_\ES$. These studies were performed on Dell Precision 3640 Intel i9 with 64 GB RAM using Python 3.9.7. 
The environment used across all case studies is shown in Fig.~\ref{fig:cs1_ts}.
The $\TS$ is shown using blue nodes and directed edges with control costs in red, or 1 otherwise. 
For this environment, we have $\vert X \vert = 20, \vert \delta \vert = 63$.

\noindent\textbf{Functionality.} Consider a team of 30 robots deployed for scientific exploration and data collection. The zones of primary interest in the environment are \textit{A}, \textit{B}, and \textit{C}. Let $S$ denote the regions to collect samples from, $M$ denotes the regions to be monitored, and $U$ denotes the data upload centers. The labels follow the notation $f_{zone,i}$ where $f$ denotes the functionality among $\{sample, monitor, upload\}$, $zone$ denotes the zone name followed by a number. For instance, for $zone A$ we have, $\{S_{A1}, S_{A2}, M_{A}\}$. The goal is to gather samples, monitor the specified regions within the given time durations, and upload the data at upload centers. 
The specification expressed plain English is:\textit{ ''Collect samples from $S_{A1}$  for 3 minutes and $S_{1}'$ for 1 minute, monitor region $M_{B1}$  for 2 minutes within the first $t_1$ minutes. Next, monitor regions $M_A$ and $M_{1}'$ for 1 minute each within the next $t_2$ minutes or collect samples and monitor regions $S_C, M_C$ for 2 and 3 minutes, respectively, within the next $t_3$ minutes. Finally, upload the collected data at the upload centers  $U_A, U_B, U_C$ within the next $t_4$ minutes."}  Using TWTL, the specification is expressed as follows: 
\begin{align*} 
\label{eq:cs1}
\phi = & [H^3 S_{A1} \land H^1 S_{A3} \land H^2 M_{B1}]^{[0,t_1]}  \\ & \cdot \left(\right. [H^1 S_{A2} \land H^1 M_A \land H^1 M_{B2}]^{[0,t_2]} \\  & \qquad \lor [H^2 S_C \land H^3 M_{C1} \land H^1 M_{C2}]^{[0,t_3]}\left.\right)  \\
   & \cdot [U_A \land U_B \land U_C]^{[0,t_4]}.
\end{align*}
Here, the infeasibilities may stem from insufficient time to reach regions or unreachable regions. In such cases, we have the relaxation preferences:
\textit{1) Region $S_{A1}$ can be substituted by collecting samples from regions $S_{1}'$ and $S_2'$ with a penalty of 5;
2) Substitute region $M_{A}$ by monitoring regions $M_1'$ and $M_{C2}$ and collecting samples from region $S_1'$ for a penalty of 2;
3) Collecting samples from region $S_{A3}$ can be substituted by monitoring regions $M_{B2}$ and $M_1'$ for a penalty of 3.}
Formally, 1) $ S_{A1} \mapsto^5 S_1' S_2'$, 2) $M_A \mapsto^2 M_1' M_{C2} S_1'$ and 3) $S_{A3} \mapsto^3 M_{B2} M_1'$.

\begin{figure*}[t]
    \centering
    \includegraphics[width=.86\linewidth]{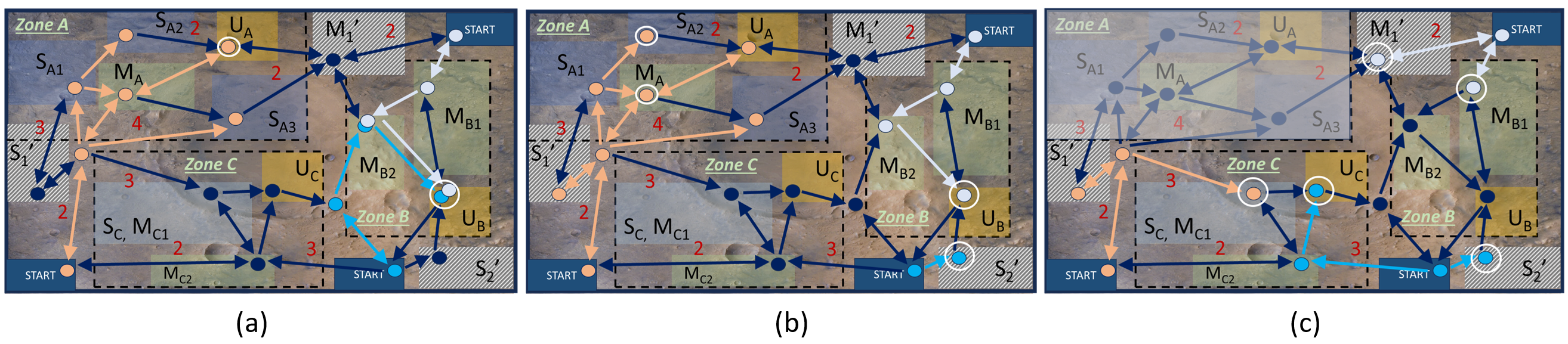}
    \caption{\small {Trajectories obtained in case of a) feasible specification, b) insufficient time, c) unavailability of $zone A$. } \vspace{-2mm}}
    \label{fig:cs1_results}
\end{figure*}

\subsubsection{Feasible specification} With $t_1 = 7, t_2 = 3, t_3 = 4, t_4 = 3$ and all regions available, $\phi$ is feasible with $\vert Q_\PA \vert = 44$, $\vert \delta_\PA \vert = 76$.
In this case, no revisions are required. Fig.~\ref{fig:cs1_results}(a) shows the generated plan where each plan starting at a different start location is shown using a different color. The final locations of each plan have been encircled. The cost is $\hat{J} = 92$ with $\lambda = 0.5$. 

\subsubsection{Insufficient time}
We assign $t_1 = 6$, keeping all other values the same as above, which makes $H^3 S_{A1}$ infeasible. Fig.~\ref{fig:cs1_results}(b) shows the plans obtained using relaxation preference (1). Note that the plan passes through $S_{A1}$ at a later time to satisfy regions $M_A, S_{A2}$.

\subsubsection{Zone unavailability} Assume zone \textit{A} is not available. Thus, relaxation preferences (1) and (3) are used and the plan is shown in Fig.~\ref{fig:cs1_results}(c). 
Even though regions $S_{A2}, M_A$ are not reachable, the subformula in disjunction is still feasible, and no relaxation is necessary for the subformula. 
Since our formulation minimizes the combined cost of control effort and relaxations, excess robots in the team are stationary.

\noindent\textbf{Efficiency Vs Expressivity.} To the authors' best knowledge, no other MILP formulation defined over timed specifications accommodates revision relaxations.
Thus, we consider a baseline MILP encoding based on flow conservation and direct specification encoding without the use of automata, similar to~\cite{sadraddini2015robust,leahy2021scalable}.
The total number of robots is N = 3.
Table~\ref{table:mtl_comparison} presents the number of binary and total decision variables and the computation time for Gurobi across six specifications.
Note that the baseline does not capture any notion of relaxation.
Thus, the number of constraints needed to guide satisfaction is considerably less.
To introduce relaxations in the baseline, additional constraints and decision variables would need to be created, increasing the computation time.
For specifications that require relaxations (e.g., last 2 specifications in Table I due to time or unavailability of $S_{A1}$), the problem is infeasible for baseline.
Thus, it is evident that MILP$_\ES$ offers more expressivity in terms of handling formal user preferences for relaxations at the cost of higher computation time. 


\begin{table}[h]
\centering
{\small
\caption{Comparison with baseline encoding}
\label{table:mtl_comparison}
\scalebox{.85}{
\begin{tabular}{|c|c|c|c|c|}
\hline
\textbf{Encoding} & \textbf{Specification} & \textbf{Binary} & \textbf{Total} & \textbf{Time (s)} \\ \hline

Baseline       &           &          10            &         484                &    0.0044        \\
MILP$_{\ES}$                   &       $[H^3 S_{A1}]^{[0,5]}$           &    1107                   &        1371              &     0.800       \\ \hline

Baseline  &  &          11            &         660                &     0.0069       \\
MILP$_{\ES}$                   & \small{$[H^2 U_C \land H^3 S_{A_2}]^{[0, 7]}$}   &         5147           &          5865               &     6.3797       \\ \hline

Baseline  &  &     7            &          649           &     0.0068     \\
MILP$_{\ES}$              &     \small{$[H^4 M_1']^{[0, 6]} \lor [H^6 M_{B1}]^{[0, 7]}$}      &   6887             &         7278            &     4.517           \\ \hline

Baseline  & &     27            &          1212          &     0.021    \\
MILP$_{\ES}$              &     \small{$[H^3 S_{A3}]^{[0, 5]} \cdot [H^2 S_1']^{[4, 9]}$}       &      28158     &         30054           &        14.07      \\ \hline

Baseline  & &                &                    &   infeasible    \\
MILP$_{\ES}$              &      \small{$[H^3 S_{A3}]^{[0, 4]} \cdot [H^2 S_1']^{[4, 9]}$}         &     25957    &      28840           &     29.382        \\ \hline

Baseline  &        \small{$[H^3 S_{A1} \land H^1 S_C]^{[0, 5]}$}  &        &                &     infeasible    \\
MILP$_{\ES}$              &   \small{$\cdot [H^2 M_{B_2}]^{[4, 8]}\cdot [H^1 U_B]^{[0,2]}$}     &   28440     &    30936   &      30.770     \\ \hline
\end{tabular}}}
\vspace{-5pt}
\end{table}
%
    \begin{figure}
        \centering
        \begin{subfigure}[b]{0.49\columnwidth}
            \centering
            \includegraphics[width=\textwidth, height=2.9cm, trim=0 0 0 -13]{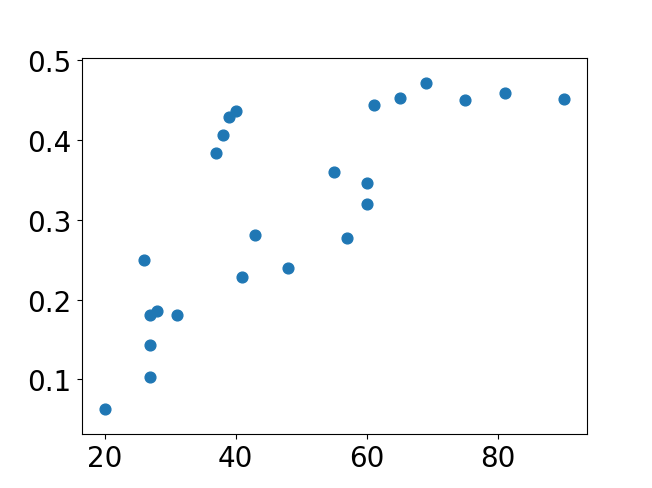}
            \caption[Network2]%
            {{\small Runtime vs $\FA$ size}}    
            \label{fig:time_dfa_nodes}
        \end{subfigure}
        \begin{subfigure}[b]{0.49\columnwidth}  
            \centering 
            \includegraphics[width=\textwidth, height=2.6cm]{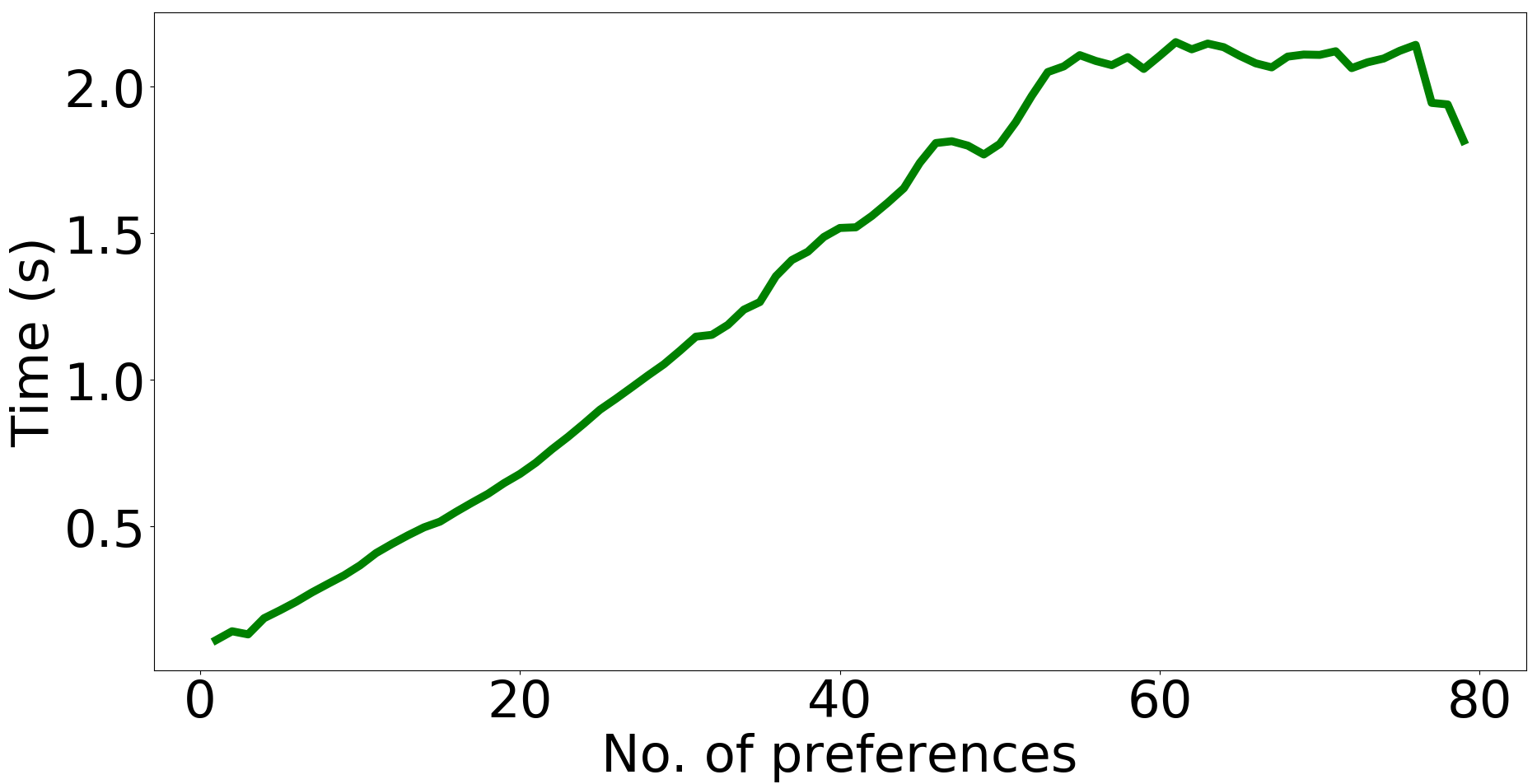}
            \caption[]%
            {{\small Runtime vs No. preferences} }   
            \label{fig:time_pref}
        \end{subfigure}
        \vskip\baselineskip
        \begin{subfigure}[b]{0.49\columnwidth}   
            \centering 
            \includegraphics[width=\textwidth, height=2.6cm]{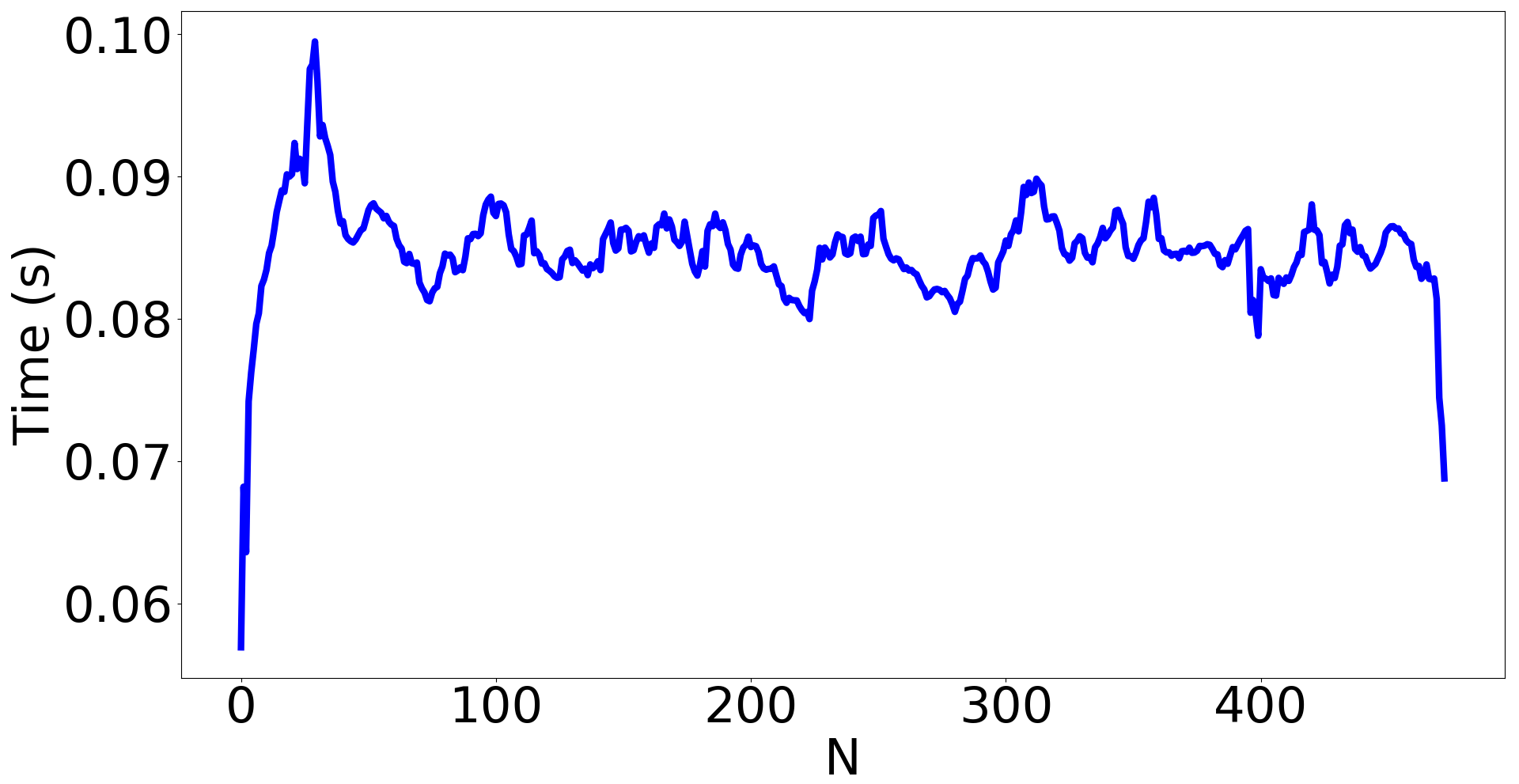}
            \caption[]%
            {{\small Runtime vs $N$}}    
            \label{fig:time_N}
        \end{subfigure}
        \begin{subfigure}[b]{0.49\columnwidth}   
            \centering 
            \includegraphics[width=\textwidth, height=2.6cm]{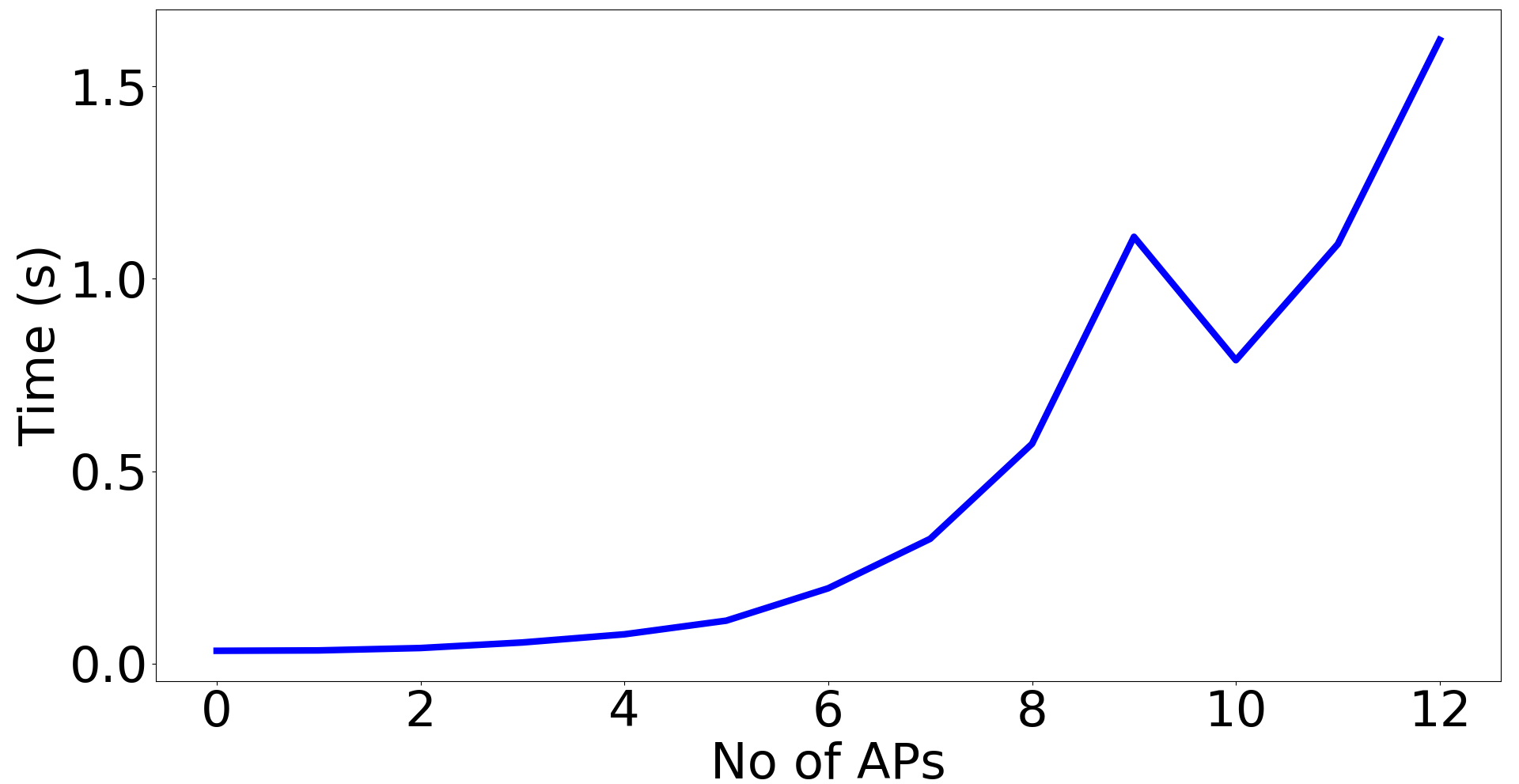}
            \caption[]%
            {{\small Runtime vs No. of APs}}    
            \label{fig:mean and std of net44}
        \end{subfigure}
        \caption[ ]
        {\small Computation time for MILP$_\ES$ for various parameters } 
        \label{fig:mean and std of nets}
    \end{figure}


\begin{figure}
    \centering
    \includegraphics[width=.9\columnwidth]{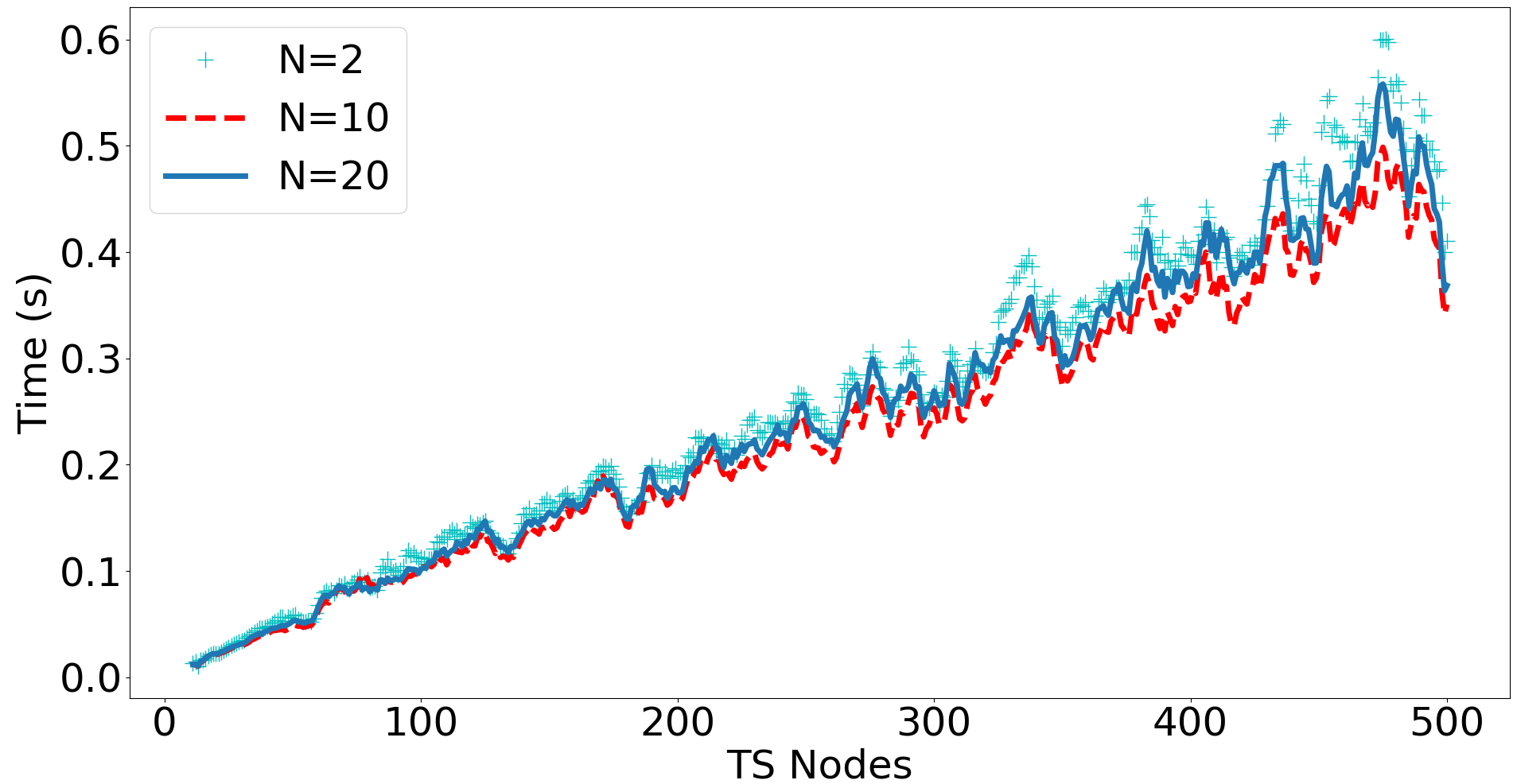}
    \caption{\small{Time for solving MILP$_\ES$ where the environment size varies from 10 to 500 for $N = 2,10,20$. }}
    \label{fig:time_ts_nodes}
\end{figure}

\subsubsection{Runtime Analysis}
We present the runtime performance of the proposed approach with respect to multiple factors, varying a single parameter at a time. Our results are in accordance with the observations made in~\cite{leahy2021scalable}. 

\noindent\textbf{$\TS$ size}. We consider $\phi = [H^1 S_{A2}]^{[0,3]} \cdot [H^2 M_A]^{[0,5]}$. The WFSE considers trivial revisions that map an input word to itself. We vary the environment size from 10 to 500 nodes, while keeping a set of APs fixed, $|AP| = 5$, from which labels are randomly assigned to the nodes. Fig.~(\ref{fig:time_ts_nodes}) shows the time performance of MILP$_\ES$ for $N = 2, 10, 20$. For smaller teams ($N=2$), the problem may take longer due to overlapping requirements. The execution time for MILPs is affected by the bounds on decision variables. Introducing more robots makes the computation faster ($N=10$), however, presence of multiple excess robots increases the upper bound on the decision variables $z_{x,q}$ and $z_{x,x', q, q'}$ resulting in a slightly higher computation time.  

\noindent\textbf{$\FA$ size}. We consider a specification of the form 
\begin{align*}
    \phi = [H^{t_1} {x_1} \land H^{t_2} {x_2}]^{[0,{t_3}]} {\small{\oplus}} [H^{t_4} {x_3}]^{[0,{t_5}]}
\end{align*}
where $\oplus \in \{\lor, \cdot \}$ and the APs $x_i$ are randomly chosen from a set \{$S_{A1}, S_{A2}, M_A, S_{A3}, S_C$\} and the durations and time windows are randomly assigned with respect to a horizon varying from 6 to 40 time units while following the conditions necessary for feasibility of $\phi$. Fig.~(\ref{fig:time_dfa_nodes}) indicates that, owing to the complex structures of $\phi$, the execution time for MILP$_\ES$ depends not only on the horizon of $\phi$ but also on its internal structure. 

\noindent\textbf{Preferences}. Given $\phi = [H^1 S_{A2}]^{[0,3]} \cdot [H^2 M_A]^{[0,5]}$ and a fixed $\TS$, we vary the number of preferences from 1 to 80 where $|AP| = 5$. We consider the preferences of the form $ B \mapsto \sigma$ where $\sigma$ is a set of propositions randomly chosen from $AP$ with varying $|\sigma| \in \range{1,4}$. Fig.~(\ref{fig:time_pref}) shows that the computation time varies almost linearly with the number of preferences and tends to be constant for a large number of preferences. Note that, similar to $\FA$, the computation for preferences also depends on their internal structure. 

\noindent\textbf{Number of robots}. Keeping the $\TS, \phi$ same as above, we vary the number of robots $N$ from 1 to 400. The results, as visualized in Fig.~(\ref{fig:time_N}), indicate that initially, there is a rise in the computation time due to an increase in the bounds of the variables. However, the computation time remains largely unaffected when $N$ increases further.  
 
\noindent\textbf{$\mathbf{AP}$ size}. Finally, we note the effect of varying the size of atomic propositions on runtime. As expected, the computation time increases exponentially with $|\AP|$ since the transitions in $\PA$ are defined over $\spow{AP}$.

\section{Conclusion}
We present a novel approach to control synthesis for multi-robot teams task with global temporal logic specifications and with user preferences for relaxations. Preserving the explicit indication of progress towards satisfaction using a relaxed specification automaton, we propose a mixed-integer linear programming approach to avoid the computationally expensive product construction of robots' motion models. The runtime analysis indicates that our approach offers more expressivity as compared to existing MILP formulations for temporal logic planning while offering computational advantages with respect to purely automata-based methods.

\bibliographystyle{ieeetr}
\bibliography{ref}
\end{document}

%% file: def.tex


\graphicspath{{figures/}}




\newtheorem{definition}{Definition}
\newcommand{\exampler}[2]{\medskip \hskip -\parindent {\bf Example #1 Revisited.~}{\it #2}\medskip}




\newcommand{\margin}[1]{\marginpar{\tiny\color{blue} #1}}
\newcommand{\todo}[1]{\vskip 0.05in \colorbox{yellow}{$\Box$ \ttfamily\bfseries\small#1}\vskip 0.05in}
\newcommand{\highlight}[2][yellow]{\mathchoice%
  {\colorbox{#1}{$\displaystyle#2$}}%
  {\colorbox{#1}{$\textstyle#2$}}%
  {\colorbox{#1}{$\scriptstyle#2$}}%
  {\colorbox{#1}{$\scriptscriptstyle#2$}}}%

\renewcommand\baselinestretch{0.98}\selectfont
\newcommand{\RM}[1]{\mathrm{#1}}
\newcommand{\CA}[1]{\mathcal{#1}}
\newcommand{\BF}[1]{\mathbf{#1}}
\newcommand{\BB}[1]{\mathbb{#1}}
\newcommand{\TT}[1]{\mathtt{#1}}
\newcommand{\BS}[1]{\boldsymbol{#1}}

\newcommand{\False}{\perp}
\newcommand{\trap}{\bowtie}
\newcommand{\virtual}{\rhd}

\newcommand{\Pref}{{R}}


\newcommand{\buchi}{B\"uchi\ }

\newcommand{\BA}{\mathcal{B}}
\newcommand{\FA}{{\mathcal{A}_\phi}}
\newcommand{\LA}{\mathcal{L}}
\newcommand{\KA}{\mathcal{K}}
\newcommand{\ES}{\mathcal{E}}

\newcommand{\ras}[1]{\stackrel{#1}{\to}}
\newcommand{\rasp}[2]{\overset{#1\mid #2}{\to}}


\newcommand{\norm}[1]{\left\| {#1} \right\|}
\newcommand{\normu}[1]{\left\| {#1} \right\|_{U}}
\newcommand{\norml}[1]{\left\| {#1} \right\|_{L}}
\newcommand{\norminf}[1]{\left\| {#1} \right\|_{\infty}}
\newcommand{\normeucl}[1]{\left\| {#1} \right\|_{2}}
\newcommand{\abs}[1]{\left| {#1} \right|}
\newcommand{\card}[1]{\left| {#1} \right|}
\newcommand{\spow}[1]{2^{#1}}
\newcommand{\lrel}[1]{\left| {#1} \right|_{LR}}
\newcommand{\indicator}{\chi}
\newcommand{\interior}[1]{\mathring{#1}}
\newcommand{\prefix}[1]{P\left({#1}\right)}
\newcommand{\dto}{\rightrightarrows}
\newcommand{\range}[1]{\left[\left[{#1}\right]\right]}
\newcommand{\any}{\bullet}
\newcommand{\R}{\mathbb{R}}

%

%% file: main.bbl
\begin{thebibliography}{10}

\bibitem{fainekos2011revising}
G.~E. Fainekos, ``Revising temporal logic specifications for motion planning,''
  in {\em 2011 IEEE international conference on robotics and automation},
  pp.~40--45, IEEE, 2011.

\bibitem{kim2012approximate}
K.~Kim and G.~E. Fainekos, ``Approximate solutions for the minimal revision
  problem of specification automata,'' in {\em 2012 IEEE/RSJ International
  Conference on Intelligent Robots and Systems}, pp.~265--271, IEEE, 2012.

\bibitem{kim2015minimal}
K.~Kim, G.~Fainekos, and S.~Sankaranarayanan, ``On the minimal revision problem
  of specification automata,'' {\em The International Journal of Robotics
  Research}, vol.~34, no.~12, pp.~1515--1535, 2015.

\bibitem{lahijanian2016specification}
M.~Lahijanian and M.~Kwiatkowska, ``Specification revision for markov decision
  processes with optimal trade-off,'' in {\em 2016 IEEE 55th Conference on
  Decision and Control (CDC)}, pp.~7411--7418, IEEE, 2016.

\bibitem{tuumova2013minimum}
J.~Tumova, L.~I.~R. Castro, S.~Karaman, E.~Frazzoli, and D.~Rus,
  ``Minimum-violation ltl planning with conflicting specifications,'' in {\em
  2013 American Control Conference}, pp.~200--205, IEEE, 2013.

\bibitem{wongpiromsarn2021minimum}
T.~Wongpiromsarn, K.~Slutsky, E.~Frazzoli, and U.~Topcu, ``Minimum-violation
  planning for autonomous systems: Theoretical and practical considerations,''
  in {\em 2021 American Control Conference (ACC)}, pp.~4866--4872, IEEE, 2021.

\bibitem{vasile2017minimum}
C.-I. Vasile, J.~Tumova, S.~Karaman, C.~Belta, and D.~Rus, ``Minimum-violation
  scltl motion planning for mobility-on-demand,'' in {\em 2017 IEEE
  International Conference on Robotics and Automation (ICRA)}, pp.~1481--1488,
  IEEE, 2017.

\bibitem{lindemann2019control}
L.~Lindemann and D.~V. Dimarogonas, ``Control barrier functions for multi-agent
  systems under conflicting local signal temporal logic tasks,'' {\em IEEE
  control systems letters}, vol.~3, no.~3, pp.~757--762, 2019.

\bibitem{vasile2014automata}
C.~I. Vasile and C.~Belta, ``An automata-theoretic approach to the vehicle
  routing problem.,'' in {\em Robotics: Science and Systems}, pp.~1--9,
  Berkeley, CA, 2014.

\bibitem{vasile2017time}
C.-I. Vasile, D.~Aksaray, and C.~Belta, ``Time window temporal logic,'' {\em
  Theoretical Computer Science}, vol.~691, pp.~27--54, 2017.

\bibitem{cardona2022partial}
G.~A. Cardona and C.-I. Vasile, ``Partial satisfaction of signal temporal logic
  specifications for coordination of multi-robot systems,'' in {\em
  International Workshop on the Algorithmic Foundations of Robotics},
  pp.~223--238, Springer, 2022.

\bibitem{cardona2023preferences}
G.~A. Cardona and C.-I. Vasile, ``Preferences on partial satisfaction using
  weighted signal temporal logic specifications,'' in {\em 2023 European
  Control Conference (ECC)}, pp.~1--6, IEEE, 2023.

\bibitem{lacerda2015optimal}
B.~Lacerda, D.~Parker, and N.~Hawes, ``Optimal policy generation for partially
  satisfiable co-safe ltl specifications.,'' in {\em IJCAI}, vol.~15,
  pp.~1587--1593, Citeseer, 2015.

\bibitem{rahmani2020you}
H.~Rahmani and J.~M. O’Kane, ``What to do when you can’t do it all:
  Temporal logic planning with soft temporal logic constraints,'' in {\em 2020
  IEEE/RSJ International Conference on Intelligent Robots and Systems (IROS)},
  pp.~6619--6626, IEEE, 2020.

\bibitem{ahlberg2019human}
S.~Ahlberg and D.~V. Dimarogonas, ``Human-in-the-loop control synthesis for
  multi-agent systems under hard and soft metric interval temporal logic
  specifications,'' in {\em 2019 IEEE 15th International Conference on
  Automation Science and Engineering (CASE)}, pp.~788--793, IEEE, 2019.

\bibitem{guo2018probabilistic}
M.~Guo and M.~M. Zavlanos, ``Probabilistic motion planning under temporal tasks
  and soft constraints,'' {\em IEEE Transactions on Automatic Control},
  vol.~63, no.~12, pp.~4051--4066, 2018.

\bibitem{kamale2021automata}
D.~Kamale, E.~Karyofylli, and C.-I. Vasile, ``Automata-based optimal planning
  with relaxed specifications,'' in {\em 2021 IEEE/RSJ International Conference
  on Intelligent Robots and Systems (IROS)}, pp.~6525--6530, IEEE, 2021.

\bibitem{kantaros2020stylus}
Y.~Kantaros and M.~M. Zavlanos, ``Stylus*: A temporal logic optimal control
  synthesis algorithm for large-scale multi-robot systems,'' {\em The
  International Journal of Robotics Research}, vol.~39, no.~7, pp.~812--836,
  2020.

\bibitem{kantaros2018sampling}
Y.~Kantaros and M.~M. Zavlanos, ``Sampling-based optimal control synthesis for
  multirobot systems under global temporal tasks,'' {\em IEEE Transactions on
  Automatic Control}, vol.~64, no.~5, pp.~1916--1931, 2018.

\bibitem{sun2022multi}
D.~Sun, J.~Chen, S.~Mitra, and C.~Fan, ``Multi-agent motion planning from
  signal temporal logic specifications,'' {\em IEEE Robotics and Automation
  Letters}, vol.~7, no.~2, pp.~3451--3458, 2022.

\bibitem{cardona2023temporal}
G.~A. Cardona, K.~Leahy, and C.-I. Vasile, ``Temporal logic swarm control with
  splitting and merging,'' in {\em 2023 IEEE International Conference on
  Robotics and Automation (ICRA)}, pp.~12423--12429, IEEE, 2023.

\bibitem{leahy2021scalable}
K.~Leahy, Z.~Serlin, C.-I. Vasile, A.~Schoer, A.~M. Jones, R.~Tron, and
  C.~Belta, ``Scalable and robust algorithms for task-based coordination from
  high-level specifications (scratches),'' {\em IEEE Transactions on Robotics},
  vol.~38, no.~4, pp.~2516--2535, 2021.

\bibitem{leahy2022fast}
K.~Leahy, A.~Jones, and C.-I. Vasile, ``Fast decomposition of temporal logic
  specifications for heterogeneous teams,'' {\em IEEE Robotics and Automation
  Letters}, vol.~7, no.~2, pp.~2297--2304, 2022.

\bibitem{gurobi}
{Gurobi Optimization, LLC}, ``{Gurobi Optimizer Reference Manual},'' 2023.

\bibitem{karlsson2020sampling}
J.~Karlsson, F.~S. Barbosa, and J.~Tumova, ``Sampling-based motion planning
  with temporal logic missions and spatial preferences,'' {\em
  IFAC-PapersOnLine}, vol.~53, no.~2, pp.~15537--15543, 2020.

\bibitem{peterson2021distributed}
R.~Peterson, A.~T. Buyukkocak, D.~Aksaray, and Y.~Yaz{\i}c{\i}o{\u{g}}lu,
  ``Distributed safe planning for satisfying minimal temporal relaxations of
  twtl specifications,'' {\em Robotics and Autonomous Systems}, vol.~142,
  p.~103801, 2021.

\bibitem{ulusoy2011optimal}
A.~Ulusoy, S.~L. Smith, X.~C. Ding, C.~Belta, and D.~Rus, ``Optimal multi-robot
  path planning with temporal logic constraints,'' in {\em 2011 IEEE/RSJ
  international conference on intelligent robots and systems}, pp.~3087--3092,
  IEEE, 2011.

\bibitem{nikou2016cooperative}
A.~Nikou, J.~Tumova, and D.~V. Dimarogonas, ``Cooperative task planning of
  multi-agent systems under timed temporal specifications,'' in {\em 2016
  American Control Conference (ACC)}, pp.~7104--7109, IEEE, 2016.

\bibitem{sadraddini2015robust}
S.~Sadraddini and C.~Belta, ``Robust temporal logic model predictive control,''
  in {\em Communication, Control, and Computing (Allerton), 2015 Annual
  Allerton Conference on}, pp.~772--779, IEEE, 2015.

\end{thebibliography}
